\documentclass[journal=jcisd8,manuscript=article]{achemso}
\setkeys{acs}{articletitle = true}

\usepackage[version=3]{mhchem} %
\usepackage{amsfonts}
\usepackage{makecell}
\usepackage{graphicx}
\usepackage{caption}
\usepackage{subcaption}
\usepackage{hyperref}
\usepackage{placeins}
\usepackage{longtable}
\usepackage{mathrsfs}
\usepackage{url}
\usepackage{todonotes}
\usepackage{mathtools}

\usepackage{color}

\author{Lior Hirschfeld}
\affiliation[MIT]{Computer Science and Artificial Intelligence Laboratory, MIT, Cambridge, MA 02139}

\author{Kyle Swanson}
\affiliation[Cambridge]{Department of Pure Mathematics and Mathematical Statistics, University of Cambridge, Cambridge, UK CB3 0WB}
\affiliation[MIT]{Computer Science and Artificial Intelligence Laboratory, MIT, Cambridge, MA 02139}

\author{Kevin Yang}
\affiliation[Berkeley]{Department of Electrical Engineering and Computer Sciences, University of California Berkeley, Berkeley, CA 94720}

\author{Regina Barzilay}
\email{regina@csail.mit.edu}
\affiliation[MIT]{Computer Science and Artificial Intelligence Laboratory, MIT, Cambridge, MA 02139}

\author{Connor W. Coley}
\email{ccoley@mit.edu}
\affiliation[MIT]{Department of Chemical Engineering, MIT, Cambridge, MA 02139}

\title{Uncertainty Quantification using Neural Networks for Molecular Property Prediction }


\begin{document}

\begin{abstract}
Uncertainty quantification (UQ) is an important component of molecular property prediction, particularly for drug discovery applications where model predictions direct experimental design and where unanticipated imprecision wastes valuable time and resources. 
The need for UQ is especially acute for neural models, which are becoming increasingly standard yet are challenging to interpret. 
While several approaches to UQ have been proposed in the literature, there is no clear consensus on the comparative performance of these models. 
In this paper, we study this question in the context of regression tasks. 
We systematically evaluate several methods on five regression datasets using multiple complementary performance metrics. 
Our experiments show that none of the methods we tested is unequivocally superior to all others, and none produce a particularly reliable ranking of errors across multiple datasets. While we believe these results show that existing UQ methods are not sufficient for all common use cases and further research is needed, we conclude with a practical recommendation as to which existing techniques seem to perform well relative to others.

\end{abstract}

\section{Introduction}
Uncertainty quantification (UQ) has grown increasingly valuable in molecular property prediction pipelines, where quantitative structure-activity relationship (QSAR) models are used to prioritize expensive and time consuming experimentation.\cite{cherkasov_qsar_2014, muratov_qsar_2020} Of the many architectures employed for QSAR modeling,  neural networks (NNs) are some of the most opaque in terms of our ability to interpret their predictions. This makes it especially challenging to evaluate their robustness, out-of-domain applicability, and possible failure modes. Significant work has been done in UQ for NNs in an attempt to resolve this problem, both within and outside of the context of QSAR\cite{kendall_what_2017,tran2019material,lakshminarayanan2016simple,cortes2019deepconfidence,gal2016dropout,cortes-ciriano_reliable_2019,ryu_uncertainty_2019,zhang_bayesian_2019,liu_molecular_2019,roy_simple_2015,aniceto_novel_2016,janet2019quantitative,huang2015scalable} including several comparative reviews\cite{toplak2014assessment,scalia2019evaluating}. 
 While many UQ methods have been developed, variation between datasets, evaluation criteria, and hyperparameter selection have made objective performance comparison difficult. Our goal is to systematically explore this topic.

We focus on small organic molecules and properties or activities used in four widely-adopted public benchmark datasets and one synthetic dataset for regression tasks.\cite{Wu_2018} We systematically evaluate a variety of UQ techniques compatible with NNs. We apply these methods to message passing networks (MPNNs), which learn parameterized mappings from graph-structured objects to continuous feature vectors and have achieved state-of-the-art performance across a wide variety of public and industrial datasets.\cite{duvenaud2015convolutional,kearnes2016molecular,gilmer2017neural,coley_convolutional_2017,Wu_2018,yang2019chemprop} In addition, we also apply these methods to feed forward networks (FFNs) using static fingerprint (FP) representations, to see if behavior is constant across NN architectures. We also test two fingerprint-based regression techniques with alternate approaches to uncertainty estimation not involving neural networks.

We explore multiple metrics for evaluating UQ that reflect its potential applications. At minimum, an effective form of UQ should help us identify molecules for which the model prediction is more or less certain than for other molecules. When the objective is to maximize the domain of applicability of the QSAR model, it could be used to select molecules with the highest uncertainty for experimental testing. To test for this desirable property, we measure each method's ability to \emph{rank} predictions by absolute error. Some UQ methods also model the expected  error as a Gaussian random variable, thereby producing explicit confidence intervals. We evaluate whether these confidence intervals reflect the distribution of observed errors; this can be applied to batched Bayesian optimization to estimate the overall probability of improvement and whether predictions are likely to be systematically over- or under-confident. We also evaluate predictions on a molecule-by-molecule basis in terms of the probability that the observed error can be explained by the quantitative uncertainty prediction. 

No method consistently satisfied the objective of producing a strong ranking of errors, on either MPNNs or FFNs. Performance of this task was strongly dataset dependent, with all methods producing poor rankings for certain tasks. We also found no UQ method to significantly outperform all others across all performance metrics. Instead, we found substantial variation in performance across all dimensions of our experimentation: Methods performed inconsistently  across evaluation criteria within each dataset;  within each evaluation criterion, methods performed inconstantly  across datasets. Our results show that while existing UQ tools may be used to isolate low-error predictions for some tasks,  researchers can still encounter tasks on which all existing methods do poorly. We have also not identified any dataset properties that make this more likely to occur, leaving trial and error as the only currently known strategy for identifying these cases. Therefore, we believe that  further method development is necessary 
to identify reliable UQ methods for regression tasks.

\section{Methods}

We distinguish between four primary strategies for uncertainty estimation: Ensemble based methods, distance based methods, mean variance estimation, and union based methods.  Some of these UQ methods are intended to provide a quantitative measure of the expected squared error, while others are intended to be used only to indicate relative uncertainty. Methods of the former type are preferred, as they can be directly incorporated into Bayesian optimization frameworks for molecular optimization. Methods of the latter type are still very useful, as they can be used for Bayesian experimental design when the goal is to optimize model accuracy. The evaluation metrics we describe later reflect both of these use cases. 

We restrict our evaluation to regression tasks, for which we have a dataset $\mathcal{D}$ comprising a collection of tuples $\{(x,y)\}$, where $y \in \mathbb{R}$ is a scalar property and $x$ is a molecule (represented by its SMILES string).

We treat our models' predictions as following a Gaussian distribution with an expected signed error of zero. The UQ values thus reflect beliefs about the \textit{unsigned} error. Specifically, the error of a model prediction $M(x)$ compared to the true value $y$ is modeled by
\begin{equation}
    M(x) - y \sim \mathcal{N}(0, \hat \sigma^2(x))  \hspace{1cm} \hat \sigma^2(x) \coloneqq f( U(x) ),
\end{equation}
where  $\hat \sigma^2(x)$ is a predicted variance that is assumed to be a simple function $f$ of any of the uncertainty metrics $U(x)$ described below. 

For those methods that attempt to provide a quantitative estimate of the variance, we assume $f$ to be the identity function so that $\hat \sigma^2(x) \coloneqq U(x)$. For relative uncertainty estimators, linear functions $f$ enable us to compute meaningful $\hat \sigma^2(x)$ values (see the discussion below of ``calibrated'' uncertainty).

\begin{figure}
    \centering
    \begin{subfigure}{230pt}
        \frame{\includegraphics[width=\linewidth]{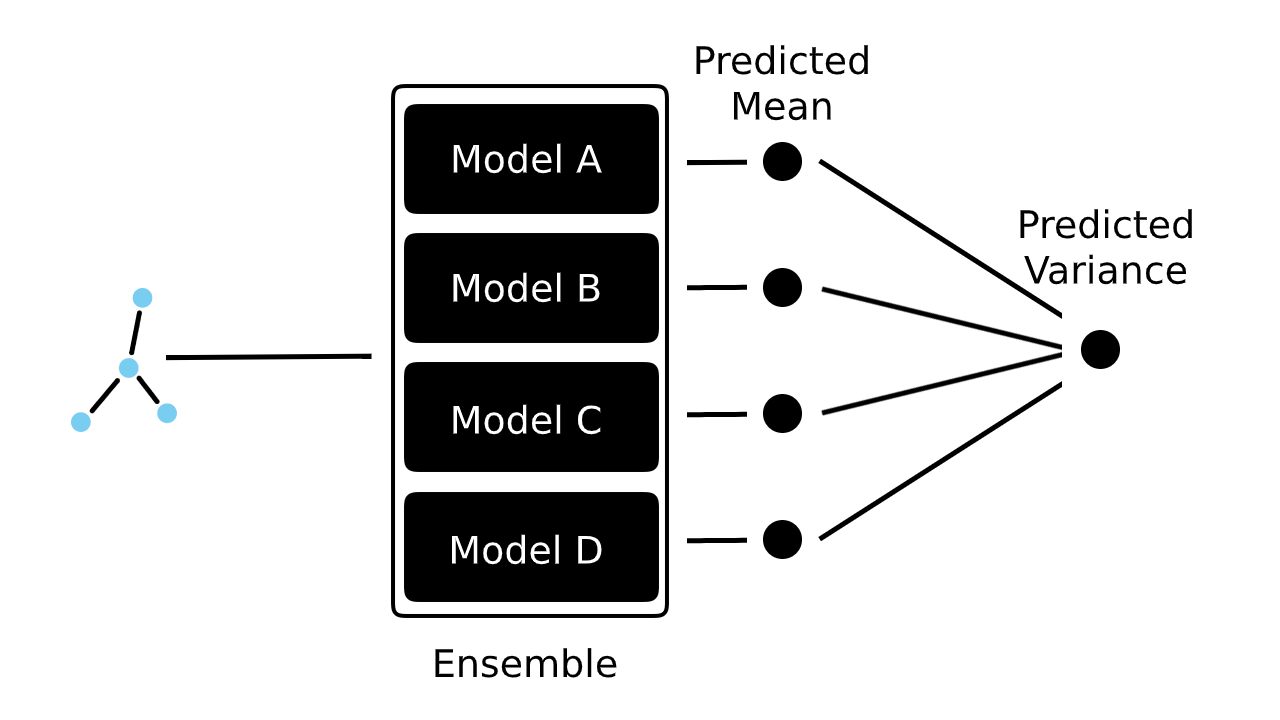}}
        \caption{Ensemble Based Methods}
        \label{fig:ensemblebased}
    \end{subfigure}
    \begin{subfigure}{230pt}
        \frame{\includegraphics[width=\linewidth]{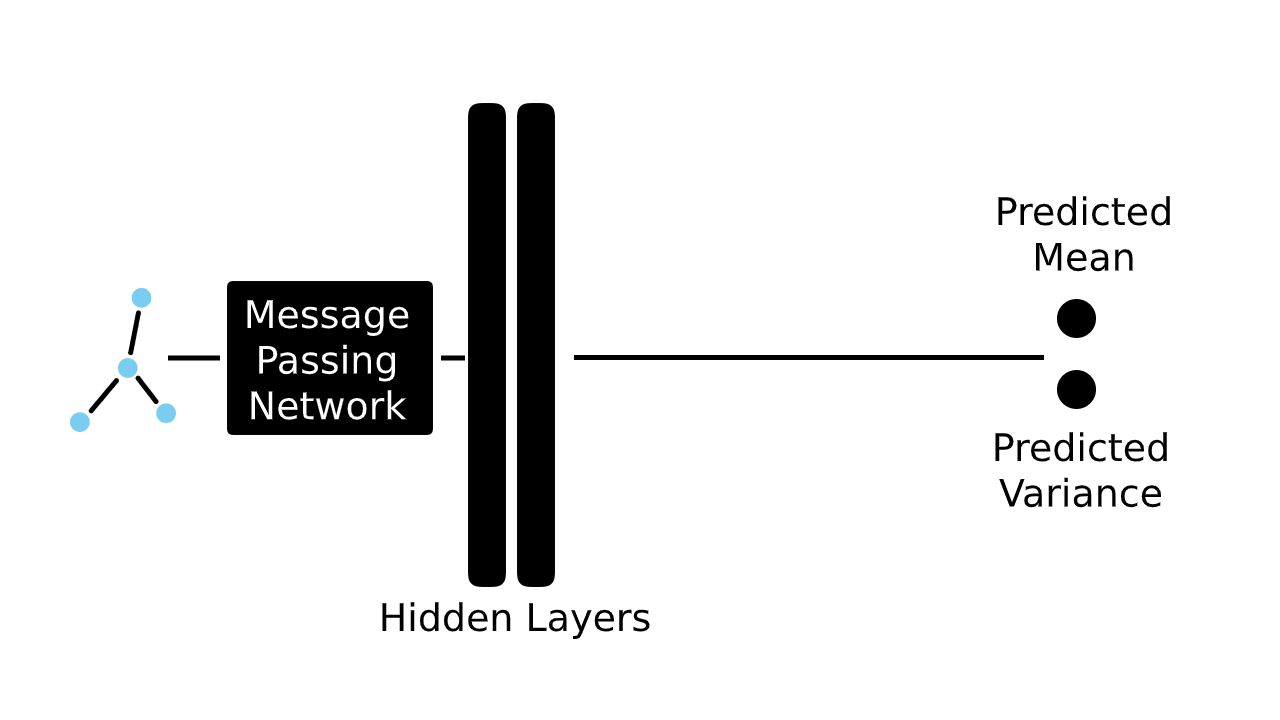}}
        \caption{Mean-Variance Estimation}
        \label{fig:meanvariance}
    \end{subfigure}

    \begin{subfigure}{230pt}
        \frame{\includegraphics[width=\linewidth]{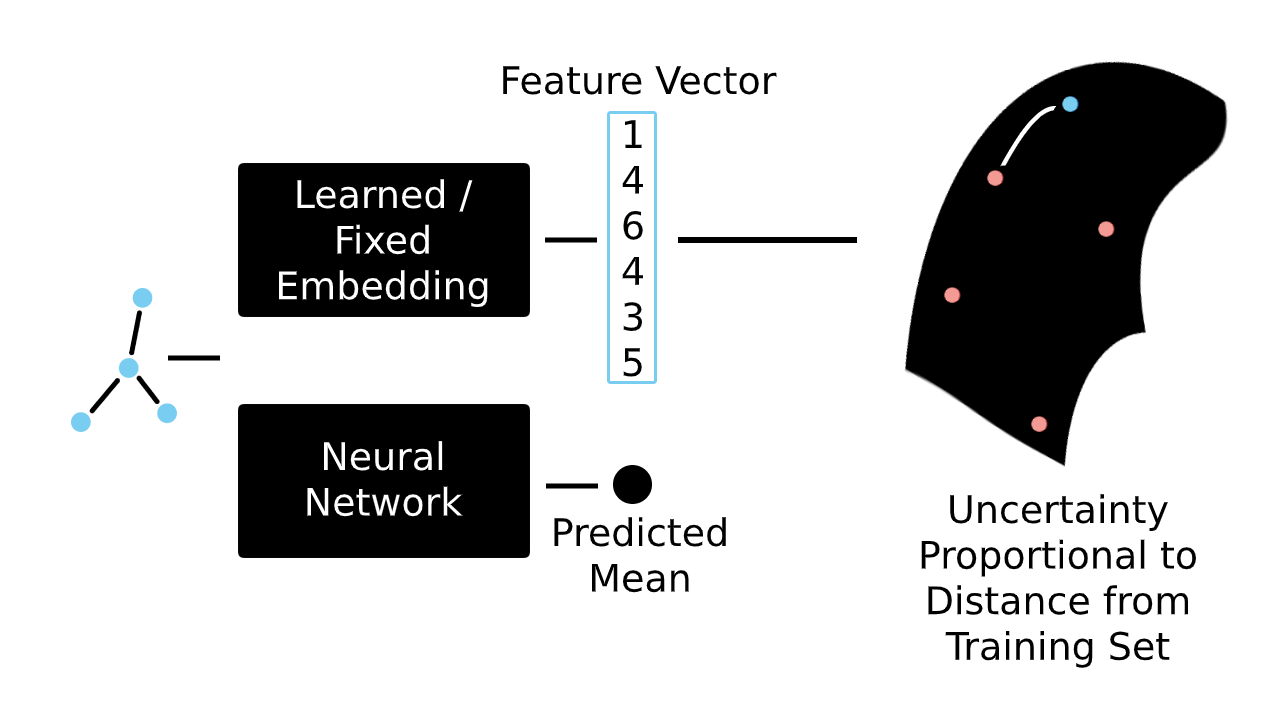}}
        \caption{Distance Based Methods}
        \label{fig:distancebased}
    \end{subfigure}
    \begin{subfigure}{230pt}
        \frame{\includegraphics[width=\linewidth]{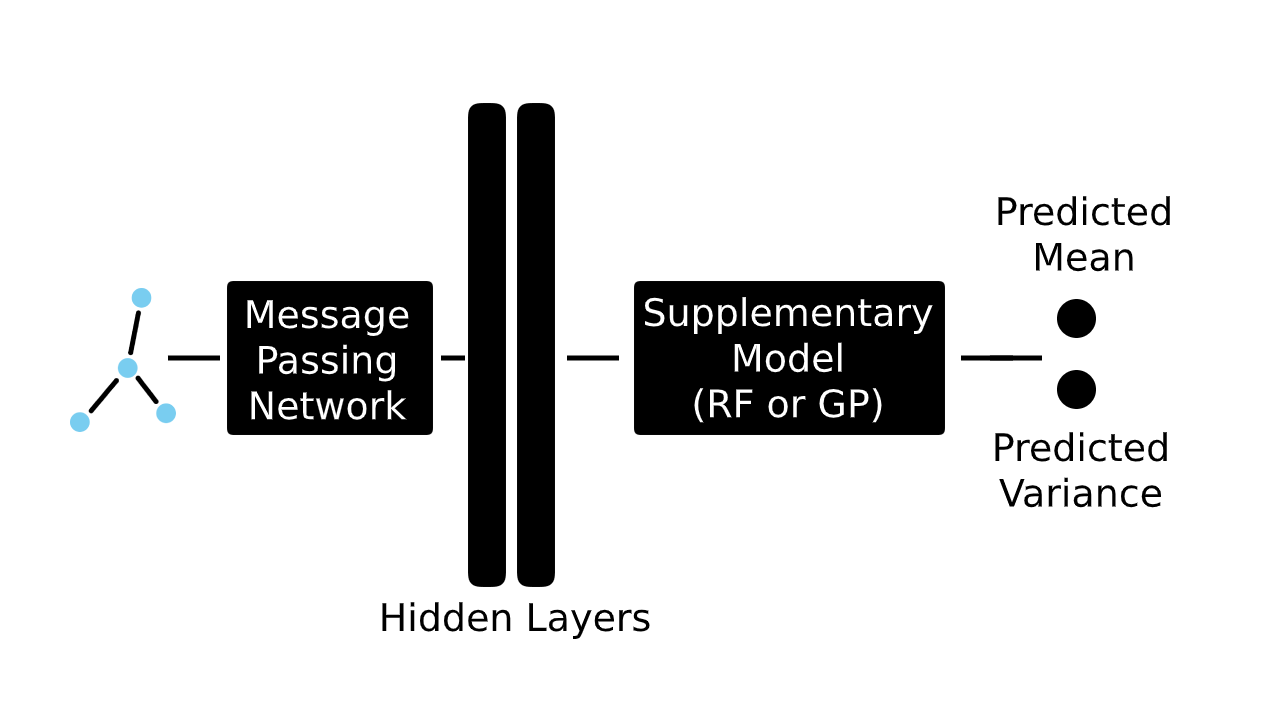}}
        \caption{Union Based Methods}
        \label{fig:unionbased}
    \end{subfigure}
    \caption{Illustration of each considered uncertainty estimator. (a) In ensemble based methods, molecules are passed through a trained ensemble, with greater variation in prediction outputs suggesting greater uncertainty. (b) In mean-variance estimation, the neural network's output is modified to produce the parameters of a Gaussian distribution. (c) In distance based methods, the distance from a molecule of interest to its nearest neighbors in the training set is interpreted as uncertainty. (d) In union based methods, the output of a neural network is fed into another model that can calculate uncertainty.}
    \label{fig:methods}
\end{figure}

The MPNN framework used in this study is Chemprop,  described in detail by \citeauthor{yang2019chemprop}.\cite{yang2019chemprop} A link to the corresponding code is available at the end of this paper. Unless otherwise noted, the model was trained with the following default parameters: $3$ message passing steps, ReLU activation functions, and a hidden size of 300 to produce a learned 300-dimensional feature vector. The 300-dimensional feature vector is passed through two additional dense layers of size 300 before a final linear output layer.

For FFNs, the same two dense layers of size 300 are used without the preceding message passing layers. For these models, molecules are represented as Morgan fingerprints as implemented in RDKit with a length of 2048, radius of 3, and the \textit{useChirality} flag set to true.\cite{landrum2006rdkit} We did not explore other FP representations. %

We now present each of the UQ strategies summarized by the four categories in Figure~\ref{fig:methods}.

\subsubsection{Ensemble Based Methods}
In any deep learning task, there is inherent stochasticity arising from a network's initialization and the order of data observed while training. For this reason, test predictions  vary  across multiple runs. Rather than training a single model $M$, ensemble based methods instead train a set of models $\mathcal{E} = \{M_1, M_2, \cdots, M_n \}$, where each $M_i$ is a single, distinct model. For any input $x$, the ensemble's prediction is defined to be the arithmetic mean of the individual models' predictions:

\begin{equation}
\tilde M(x) = \sum_{M \in \mathcal{E}} \frac{M(x)}{n}.
\end{equation}

Prior work has established that ensembling can yield models with greater accuracy than any individual model in the set \cite{dietterich2000ensemble}. \citeauthor{lakshminarayanan2016simple} propose that the variance between model outputs might also be used as a measurement of uncertainty.\cite{lakshminarayanan2016simple} Therefore, the uncertainty of an ensemble, $U_\text{ensemble}$, is defined as the variance of predictions:

\begin{equation}
U_\text{ensemble}(x) = \sum_{M \in \mathcal{E}} \frac{(\tilde M(x) - M (x))^2}{n}.
\end{equation}

The primary drawback of any ensemble based uncertainty estimation method is increased training time, the magnitude of which depends  on the size of the ensemble and how elements are selected. For testing purposes, we fix the size of the ensemble to $16$ and consider the following methods to produce $\mathcal{E}$:

\begin{description}
    \item[Traditional Ensembling]
    \hfill \\
    Each $M_i \in \mathcal{E}$ is trained on the same training data but is initialized with a different set of randomly selected weights, as proposed by \citeauthor{lakshminarayanan2016simple}.\cite{lakshminarayanan2016simple} The expectation is that in regions of the input space that are not well covered by the training data, each model is affected more significantly by its initialization, while near known values each model  is likely to have converged to a similar value. This would imply that when variance between model outputs is high, there is greater uncertainty in the ensemble's prediction. As each $M_i$ is trained individually, the computational cost scales linearly with the size of the ensemble, far exceeding the cost of training a single network.
    
    \item[Bootstrapping]
    \hfill \\
    Motivated by bootstrapping, a classic method of estimating statistical uncertainty, bootstrap ensembling trains independent models on different training data subsets and has been observed to outperform traditional ensembling for deep learning models by some metrics.\cite{scalia2019evaluating} As in traditional ensembling, each $M_i \in \mathcal{E}$ is trained independently, but now each model is only trained on a random subset of the data $\mathcal{X}_i \subset \mathcal{D}_\text{train}$. By training each model in the ensemble on different data, the ensemble gains information regarding the density of different features within our input space: It is expected that several models lack exposure to those features which are sparse in the training set and so produce high variation between model outputs. If $|\mathcal{X}_i| \ll |\mathcal{D}_\text{train}|$, bootstrapping can greatly reduce training time when compared to traditional ensembling. To test this method, we expose each $M_i$ to $25\%$ of the molecules in the training set, selected independently at random.
    \item[Snapshot Ensembling]
    \hfill \\
    This method works by training a single model $M$ and periodically storing \textit{snapshots} of its state throughout training. It is these snapshots which collectively make up the ensemble. Since training only needs to be performed once, this strategy results in significantly lower training times than traditional or bootstrap ensembling. This approach was formalized by \citeauthor{cortes-ciriano_reliable_2019}, who showed its performance to be on-par with traditional ensembling on many drug discovery tasks.\cite{cortes-ciriano_reliable_2019} To test this method, we produce snapshots by saving the weights of our model every third training epoch. After each snapshot is taken, the model's learning rate is reset in order to encourage variation between models and prevent the entire collection from converging to the same local minimum.
    \item[Monte Carlo Dropout Ensembling]
    \hfill \\
    This method begins by training a single model $M$ with dropout, stochastically setting the weight of each node in the neural network to zero at every training step with probability $p$. Dropout was originally introduced as a form of regularization to avoid overfitting, but here dropout is also applied when predicting: Rather than producing a fixed ensemble of models $\mathcal{E}$, a set of predictions is produced for every input $x$ by applying different randomly generated dropout masks. \citeauthor{gal2016dropout} showed that such a strategy approximates Bayesian inference.\cite{gal2016dropout} Monte Carlo dropout ensembling is advantageous in that it significantly cuts down on computational cost by only training a single instance of the model. This approach has been applied to QSAR tasks, e.g., by \citeauthor{cortes-ciriano_reliable_2019}.\cite{cortes-ciriano_reliable_2019} 
    Following \citeauthor{gal2016dropout}, we test this strategy's performance with dropout rates of $10\%$ and $20\%$.\cite{gal2016dropout} 
\end{description}

\subsubsection{Mean Variance Estimation (MVE)}

Ideally, a neural network would be capable of estimating the uncertainties of its own predictions. One  framework with this capability is the \textit{Bayesian neural network}, in which every weight parameter is represented by a Gaussian random variable, so the final prediction is described by a distribution, rather than a fixed value.\cite{ryu_uncertainty_2019,zhang_bayesian_2019,carr_graph_2019} However, training these networks using an exact loss function is impractical for large datasets.%

For this reason, it is useful to consider alternate distributional approaches like Mean Variance Estimation (MVE).\cite{nix_estimating_1994} In this approach, the output layer of the network is modified to predict both the mean $\mu(x)$ and variance $\sigma^2(x)$ of the property of interest for input $x$; the softplus activation is used to ensure $\sigma^2(x)$ is strictly positive. The model is trained using a negative log likelihood loss according to a Gaussian likelihood model, defined as:
\begin{equation}
    \text{NLL}(x,y) = \frac{1}{2}\ln (2 \pi) + \frac{1}{2}\ln (\sigma^2(x)) + \frac{(y - \mu(x))^2}{2\sigma^2(x)}. 
\end{equation}
The network's predicted variance, $\sigma^2(x)$, is then treated as the uncertainty. Alternate implementations of this concept have trained neural networks to predict the logarithm of the variance rather than the variance itself.\cite{kendall_what_2017}

\subsubsection{Distance Based Methods}
One intuitive reason to be uncertain about predictions made for certain test inputs is if they are  dissimilar from inputs in the training set. For a particular input $x$ and distance measurement $d(\cdot, \cdot)$, the uncertainty of the model $M$ on $x$ can be defined as the minimum distance between $x$ and any element in the training set:

\begin{equation}
U_\text{dist}(x) = \min \{d(x, x^\prime) : x^\prime \in \mathcal{D}_\text{train} \}.
\end{equation}

Quantitative measurements of distance between molecules could identify molecules that are far from the training set and, ostensibly, more likely to produce high error.\cite{sheridan_similarity_2004,liu_molecular_2019} However, as this calculation is highly sensitive to outliers, a more robust uncertainty estimate is to instead consider $U_\text{dist}(x)$ to be the average distance between $x$ and its $k$ nearest neighbors in the training set. In our experiments, we set $k=8$. Distance metrics implicitly assume that there is a relatively smooth mapping between the input representation and the output value, so it is worth noting that this assumption is clearly violated for some molecular property prediction tasks due to the known presence of structure-activity cliffs.\cite{stumpfe_exploring_2012} 

What remains then is to identify a distance measurement for which this assumption holds. We consider two principle strategies:

\begin{description}
    \item[Structure Space]
    \hfill \\
     One candidate is the log-scaled Tanimoto distance between two binary molecular fingerprints $x, x^\prime$:

    \begin{equation}
        D_T(x, x^\prime)  = -\log _{2} \left (\frac{|x \cap x^\prime|}{|x \cup x^\prime|} \right ).
    \end{equation}
    
    The Tanimoto distance provides a metric of relative uncertainty, not a metric that is intended to match the magnitude of error. To encode molecules, we use Morgan fingerprints as implemented in RDKit with a length of 2048, radius of 3, and the \textit{useChirality} flag set to true. 
    
    There are many descriptors and metrics that could be used to assess distance between molecules, including functional descriptors and the Mahalanobis distance, as used by \citeauthor{toplak2014assessment} and \citeauthor{schroeter2007estimating}.\cite{toplak2014assessment,schroeter2007estimating} Later work proposed identifying molecules outside of the applicability domain of a QSAR model (i.e., a binary assessment of uncertainty) solely based on the values of each descriptor relative to the distribution observed in the training set.\cite{roy_simple_2015} There are also ways of incorporating information about the predicted reliability of neighbors based on the density of compounds in chemical space, rather than using a strict nearest neighbors approach.\cite{aniceto_novel_2016} 
    
    \item[Latent Space]
    \hfill \\
    One downside of using distance in structure space for UQ is that any hardcoded measurement will not vary between prediction tasks. An alternative is to calculate distance using a learned molecular embedding from the model, as the ability to learn task-specific representations without manual feature engineering is one of the advantages of deep learning. For testing purposes, we use the Euclidean distance in a continuous embedding space--the vector representation of a molecule immediately before the output layer--to measure dissimilarity, as proposed by \citeauthor{janet2019quantitative}.\cite{janet2019quantitative} Using the default Chemprop hyperparameters, this corresponds to a 300-dimensional feature vector. 
\end{description}

Note that results are reported for both MPNN Tanimoto Distance and FFN Tanimoto Distance. While  UQ in each case is based solely on the fixed molecular fingerprint and will be identical, the predicted means will differ between the two base models; in turn, the true errors and evaluation metrics will differ.

\subsubsection{Union Based Methods}

Another method for confidence estimation, as proposed by \citeauthor{huang2015scalable}, is to combine models that naturally produce confidence estimates with the predictive power of neural networks in a pipelined approach.

In this method, a neural network $M$ is first trained for the regression problem as normal. The output layer of the network is then dropped to provide an embedding network, $M^\prime$, that transforms any input $x$ into the task-specific latent space. The entire validation set is then passed through this model to produce
\begin{equation}
\mathcal{D}_\text{val}^\prime = \{(M^\prime(x),y) : (x,y) \in \mathcal{D}_\text{val} \}.
\end{equation}

$\mathcal{D}_\text{val}^\prime$ is then used to train a second model which is capable of predicting both the regression value and the uncertainty. We consider two specific cases:

\begin{description}
    \item[Gaussian Process]
    \hfill \\
    The transformed validation set is used to train a linear Gaussian process (GP), $G$. On any further input $x$, we define our prediction and uncertainty to be the output mean and variance, respectively, of $G(M^\prime(x))$. \citeauthor{huang2015scalable} show that these ``DNN-GP''s are significantly more accurate than traditional Gaussian processes.\cite{huang2015scalable} For our testing, we use  \textbf{GPy}'s \textit{SparseGPRegression} implementation, as training time with the standard \textit{GPRegression} implementation was impractical, coupled with a linear kernel.\cite{gpy2014}
    \item[Random Forest]
    \hfill \\
    The transformed validation set is used to train a random forest (RF), $R$. On any further input $x$, we take the prediction to be $R(M^\prime(x))$ and the uncertainty to be the variance of predictions of the decision trees. For our testing, we use  \textbf{sklearn}'s \textit{RandomForestRegressor} with $128$ trees.\cite{pedregosa2011scikit}
\end{description}

In order to train an accurate Gaussian process or random forest, a substantial amount of data must be reserved for the validation set. $\mathcal{D}_\text{val}$ is transformed to train the supplementary model rather than $\mathcal{D}_\text{train}$ because any overfitting of the neural network on the selected set would expose the supplementary model to an unrepresentative distribution of errors. This same $\mathcal{D}_\text{val}$ is used for early stopping during training as normal.

\subsubsection{Fingerprint-based Methods}
All of the methods described earlier in this section are implemented for the MPNN-based and FFN-based regressions. Also included in our evaluation are two baseline approaches that use static FP representations:
\begin{description}
    \item[FP GP]
    \hfill \\
    Gaussian Process Modeling using the fixed fingerprint in place of the learned feature vector. As with the union based GP approach, the prediction and uncertainty are taken to be the mean and variance of the GP.
    \item[FP RF]
    \hfill \\
    Random Forest modeling using the fixed fingerprint in place of the learned feature vector. As with the union based RF approach, the FP RF prediction and uncertainty are taken to be the mean and variance of the predictions of $128$ decision trees.
\end{description}

\subsection{Evaluation Metrics}

\subsubsection{Spearman's Rank Correlation Coefficient}
For any uncertainty estimator, we expect that predictions with low uncertainty will have comparatively low true error. More precisely, given some model $M$ and two molecules $a$ and $b$ for which $U(a) < U(b)$,  we  expect $M(a)$ to be more accurate than $M(b)$ \emph{on average}. This property can be measured quantitatively using Spearman's Rank Correlation Coefficient. Given two vectors $L_1$ and $L_2$, define rank vectors $r_{L_1}$ and $r_{L_2}$ that assign integer ranks to each value in ascending order. The correlation coefficient is then defined as function of their covariance (cov) and standard deviations (std), where
\begin{equation}
    \rho(L_1, L_2) = \frac{\mathrm{cov}(r_{L_1}, r_{L_2})}{\text{std}(r_{L_1}) \text{std}(r_{L_2})}.
\end{equation}
In the event that $L_1$ and $L_2$ have the exact same ranking, $\rho(L_1, L_2) = 1$, and in the event that $L_1$ and $L_2$ have opposite rankings, $\rho(L_1, L_2) = -1$. 

The ranking coefficient is calculated between the list of predicted uncertainties and the list of absolute errors. Note that a perfect correlation of $\rho=1$ is not expected since we believe the errors will be approximately normally distributed. It is possible for the model to randomly produce a result with low error even in the event it has high uncertainty. Because this metric compares ranks, not precise values, it is appropriate for relative UQ metrics (e.g., the distance based methods described above). %

\subsubsection{Miscalibration Area}
Another way to evaluate UQ methods is to consider whether predicted uncertainties are similar in magnitude, quantitatively, to the true errors we observe. This property is referred to as \textit{calibration}. \citeauthor{tran2019material} propose to compare the observed fraction of errors falling within $z$ standard deviations of the mean to what is expected for a Gaussian random variable with variance equal to the uncertainty prediction $U(x)$ of the UQ method.\cite{tran2019material} For example, 68\% of the test errors are expected to be within one standard deviation, 95\% to be within two, etc.

Given a perfect uncertainty estimator, the expected fraction should exactly match the observed fraction for any $z$. To provide a quantitative statistic, the miscalibration area is computed as the area between the true curve of observed versus expected fractions and the parity line; a perfect UQ metric will have an area of $0$. This metric only provides an  evaluation of whether a method is \textit{systematically} overconfident or \textit{systematically} underconfident. A method that is overconfident about half of the points and underconfident about the other half can still achieve a perfect score of $0$.

\subsubsection{Negative Log Likelihood (NLL)}
\label{sec:nll}
A third metric is the negative log likelihood (NLL) of observed errors under the assumption that they are normally distributed around 0 with variances given by the UQ estimates, $U(x)$. \cite{lakshminarayanan2016simple} Given some collection of molecules $\mathcal{D}_\text{test}$, the average negative log likelihood is defined to be

\begin{equation}
    NLL(\mathcal{D}_\text{test}) = \frac{1}{2|\mathcal{D}_\text{test}|} \sum _{x, y \in \mathcal{D}_\text{test}} \ln (2 \pi) + \ln (U(x)) + \frac{(M(x)-y)^2}{U(x)},
\end{equation}
where $M(x)$ is the model's point prediction and $U(x)$ is the UQ estimate for molecule $x$. Note that NLL is averaged to prevent bias against larger datasets.

For a given list of residuals, there will be a lower bound to the NLL that can be achieved. Therefore, to decouple UQ performance from the accuracy of predictions, we also examine the difference between the minimum possible NLL and the NLL that is achieved. These results can be found in the Supporting Information (Figure~\ref{fig:nll_difference}).

\subsubsection{Calibrated NLL}

For UQ methods whose uncertainty estimates are not intended to be used as variances (e.g., distances), NLL as described above is an uninformative metric. In order to still apply NLL to these methods, an alternative is to first calibrate the uncertainty estimates so that they more closely resemble variances. Thus, rather than taking $\hat \sigma^2 (x) \coloneqq U(x)$, the two are assumed to be linearly related:
\begin{equation}
    \hat \sigma^2(x) \coloneqq a U(x) + b
\end{equation}
For each dataset and method, the Calibrated NLL (cNLL) is computed using the scalars $a$ and $b$ which minimize the NLL of errors in the validation set:

\begin{equation}
    cNLL(\mathcal{D}_\text{test}) \coloneqq  \frac{1}{2 |\mathcal{D}_\text{test}|} \sum _{x, y \in \mathcal{D}_\text{test}} \ln(2 \pi) + \ln (a_* U(x) + b_*) + \frac{(M(x)-y)^2}{a_* U(x) + b_*},
\label{eq:cnll}
\end{equation}
\begin{equation}
    a_*,b_* = \underset{a,b}{\mathrm{argmin}} ~~\frac{1}{2} \sum _{x, y \in \mathcal{D}_\text{val}} \ln(2 \pi) + \ln (a U(x) + b) + \frac{(M(x)-y)^2}{a U(x) + b}.
\end{equation}
A similar recalibration was performed by \citeauthor{janet2019quantitative} in their implementation of the latent space distance metric.\cite{janet2019quantitative}. A plot of the difference between the minimum possible cNLL and the cNLL that is achieved can also be found in the Supporting Information (Figure~\ref{fig:cnll_difference}).

\subsection{Data}

We use four datasets commonly used for benchmarking from \citeauthor{Wu_2018}\cite{Wu_2018}: aqueous solubility (Delaney), solvation energy (freesolv), lipophilicity (lipo), and atomization energy (QM7). We also include one additional synthetic dataset of CLogP as a case where there is no aleatoric uncertainty--originating from inherent, unpredictable noise in the data, including experimental noise when the data were acquired\cite{kendall_what_2017}--and the property is calculable through a simple heuristic function. The CLogP dataset was prepared by taking the union of molecules appearing in the four other datasets and using the RDKit\cite{landrum2006rdkit} implementation of the Crippen heuristic estimate of the octanol-water partition coefficient.\cite{wildman_prediction_1999}

We use a 50/20/30 split for training/validation/testing. Several of the methods we evaluate (i.e., the union based methods) require the use of a large validation set to produce strong confidence estimations. All methods were tested with the same set of eight random splits to control for random performance variations resulting from a particularly lucky or unlucky test set. We additionally use a single scaffold split using RDKit's Murcko scaffold decomposition and a greedy bin packing algorithm to approximately distribute molecules according to a 50/20/30 split after sorting clusters of examples by descending cluster size; in this manner, test molecules are more likely to be  ``out-of-domain'' and have rarer scaffolds than in a random split.\cite{landrum2006rdkit}

\section{Results}

In the following subsections, we show and discuss model performance according to each of the four evaluation metrics described above. We conclude with a pairwise comparison between all models to summarize the results across all five datasets and eight random splits.

\subsubsection{Spearman's Rank Correlation Coefficient}
\begin{figure}
    \centering
        \includegraphics[scale=0.5]{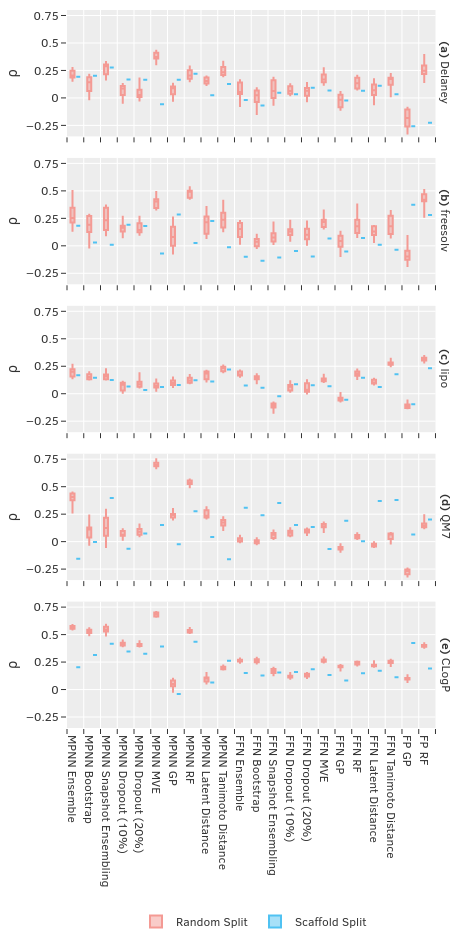}
    \caption{Spearman's rank correlation coefficient, $\rho$, for all UQ methods and datasets evaluated in this work. A higher value of $\rho$ indicates stronger agreement between the estimated uncertainty and the relative magnitude of the observed absolute error. (a) Delaney aqueous solubility, (b) freesolv solvation energy, (c) lipophilicity, (d) QM7 atomization energy, (e) CLogP heuristically-calculated lipophilicity. Random split boxes show quartiles without excluding outliers.}
    \label{fig:spearman}
\end{figure}

Figure~\ref{fig:spearman} shows the Spearman's rank correlation for all datasets and all methods. An immediate observation is that there is a significant amount of variation both between and within datasets. For some (e.g., lipophilicity in Figure~\ref{fig:spearman}c), no method is able to perform particularly well. For this experimental dataset, the measurement noise/error during acquisition can be high, on the order of 0.5 log10 unit, and the range of measurements is quite narrow; we should not necessarily expect a method to be able to perform well. For other datasets, many methods perform quite well (e.g., the simulated CLogP data in Figure~\ref{fig:spearman}e). The variation of performance across different datasets makes it clear that if one were to benchmark UQ methods on a single dataset, one could draw completely incorrect conclusions about the  superiority of a particular estimator. In fact, no method was consistently strong across all datasets, although MPNN MVE, MPNN RF, and FP RF are often highly ranked. Occasionally, a method will even exhibit a negative Spearman correlation.

\begin{figure}
    \centering
        \includegraphics[scale=0.5]{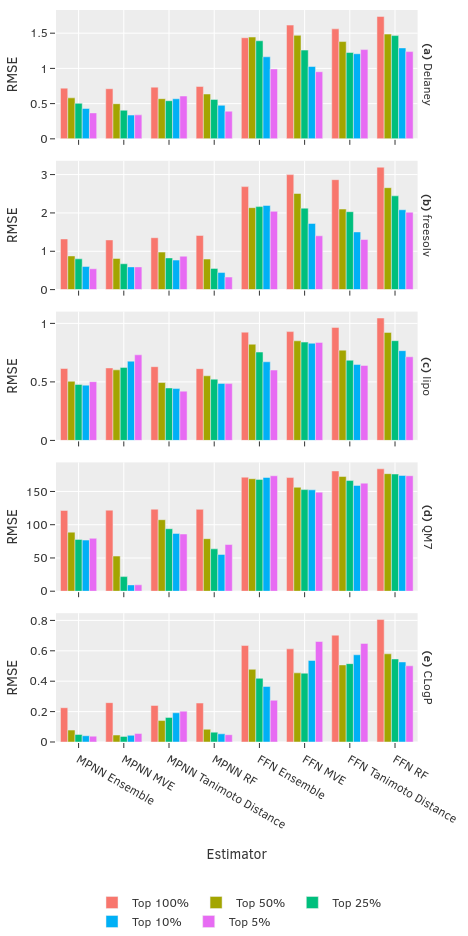}
    \caption{Average RMSE in random splits calculated for all datasets evaluated in this work and a selection of UQ methods. RMSE is recorded for the 100\%, 50\%, 25\%, 10\%, and 5\% of the test set on which the UQ method calculated the lowest uncertainty.  The UQ methods above were selected because they produced the highest median Spearman's rank correlation coefficient across datasets within the four categories of approaches summarized in Figure~\ref{fig:methods}. Datasets are (a) Delaney aqueous solubility, (b) freesolv solvation energy, (c) lipophilicity, (d) QM7 atomization energy, (e) CLogP heuristically-calculated lipophilicity.}
    \label{fig:rmse}
\end{figure}

To illustrate this phenomenon, Figure~\ref{fig:rmse} shows the change in RMSE as prediction is restricted to subsets of $\mathcal{D}_{test}$ on which the model is least uncertain. A sharp, monotonic decrease in RMSE between these classifications would indicate that the UQ correctly identifies its lowest error predictions. When employed on the lipophilicity data set, MPNN MVE actually produced higher than average error on the predictions it was most confident about, despite providing more reliable error rankings on other datasets. 

We also note that with few exceptions, scaffold splitting resulted in a slightly weaker ranking than the median random split.

\subsubsection{Miscalibration Area}

\begin{figure}
    \centering
        \includegraphics[scale=0.5]{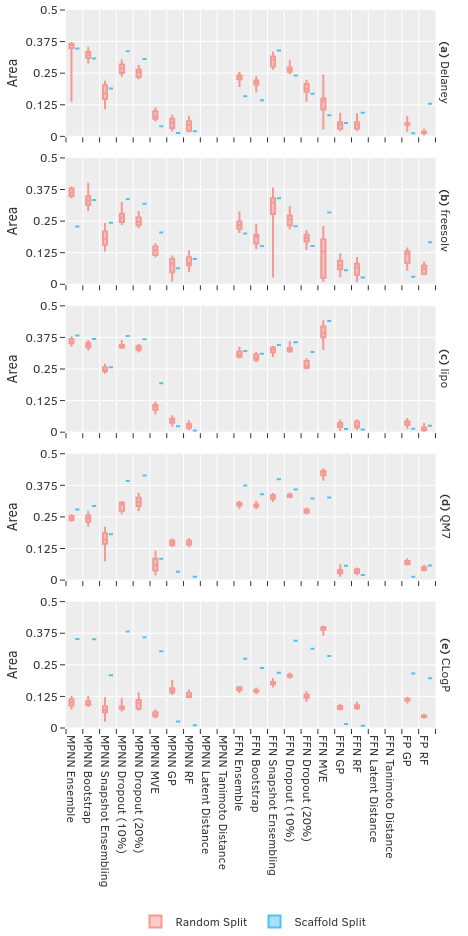}
    \caption{Miscalibration area (Area) for all UQ methods and datasets evaluated in this work. A miscalibration area closer to zero indicates better quantitative agreement between the expected and observed fraction of true values within their predicted confidence intervals at any level of statistical significance. This evaluation interprets the UQ metric quantitatively, so we exclude distance based metrics (in latent space or in structure space) which would not produce meaningful results.  (a) Delaney aqueous solubility, (b) freesolv solvation energy, (c) lipophilicity, (d) QM7 atomization energy, (e) CLogP heuristically-calculated lipophilicity.}
    \label{fig:miscalibration}
\end{figure}

Figure~\ref{fig:miscalibration} shows the miscalibration area (Area) for all datasets and all methods. Recall that miscalibration area measures systematic over- or under-confidence in an aggregated, quantitative sense. A score of $0$ indicates perfect alignment, while a  score of $0.5$ indicates maximal misalignment. As such, it does not measure if the absolute errors of \emph{individual compounds} are predicted well or poorly, nor is it appropriate to evaluate relative UQ metrics like the Tanimoto or latent space distances. 

The MPNN MVE and union methods (MPNN-/FFN-based RF/GP) are superior to most other methods, although FP GP and FP RF also perform quite well. Unlike with Spearman's $\rho$, there is a more noticeable difference between performance on random and scaffold splits. For example, Figure~\ref{fig:miscalibration}e shows that the first six MPNN methods have miscalibration areas several times higher for the scaffold split than for any random split. This is consistent with \citeauthor{scalia2019evaluating}'s observation that epistemic uncertainty--arising from either a poor model fit or a lack of exposure in the training set\cite{kendall_what_2017}--is consistently underestimated using ensembling or dropout techniques for out-of-domain samples;\cite{scalia2019evaluating} note that recalibrating the variance of an ensemble prediction to provide more reliable confidence intervals can reduce miscalibration, though we do not investigate recalibrated miscalibration areas in this study.\cite{cortes-ciriano_reliable_2019} The MPNN GP and MPNN RF approaches work remarkably well for the scaffold split, often achieving a lower miscalibration area than for the random split.

\subsubsection{Negative Log Likelihood (NLL)}
\label{sec:nll}

\begin{figure}
    \centering
        \includegraphics[scale=0.5]{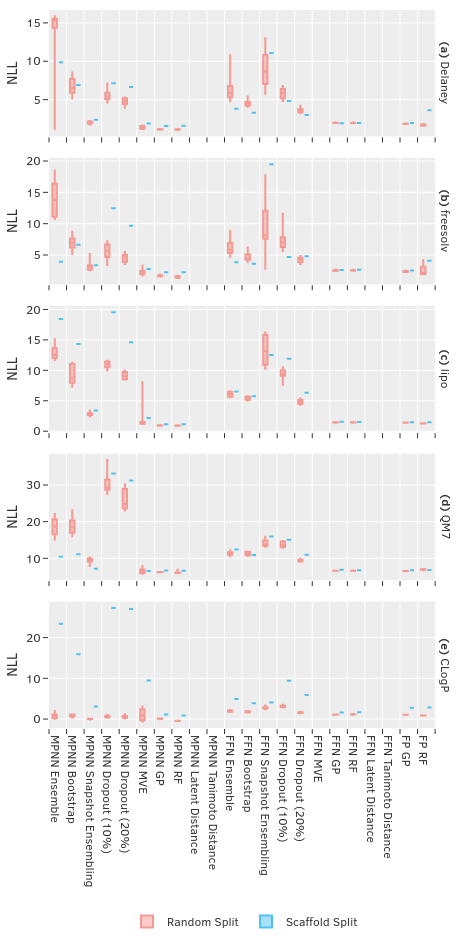}
    \caption{Negative log likelihood (NLL) for all UQ methods  (excluding  FFN MVE as an outlier) and all datasets evaluated in this work. A lower value of NLL indicates better quantitative agreement between the estimated uncertainty and the magnitude of the observed absolute error when treating the UQ metric as the predicted variance of errors. This evaluation interprets the UQ metric quantitatively, so we also exclude distance based metrics (in latent space or in structure space) which would not produce meaningful results. (a) Delaney aqueous solubility, (b) freesolv solvation energy, (c) lipophilicity, (d) QM7 atomization energy, (e) CLogP heuristically-calculated lipophilicity. Random split boxes show quartiles without excluding  outliers.}
    \label{fig:nll}
\end{figure}

The negative log likelihood (NLL) results (Figure~\ref{fig:nll}) represent a stricter evaluation of whether models are capable of predicting accurate UQ metrics that can be directly treated as variances of normally-distributed errors. It is important to keep in mind the accuracy of each method will have an affect on the minimum NLL that can be achieved through optimal prediction of variance. The more accurate a method is, the lower its errors will tend to be, and the higher its maximum likelihood, corresponding to a lower minimum NLL. A comparable figure showing the difference between observed NLL and optimal NLL can be found in Figure~\ref{fig:nll_difference}. Again, it is inappropriate to calculate NLL for relative UQ metrics, so values for Tanimoto and latent space distances are excluded. FFN MVE also produced such extreme values on certain splits that it has been omitted for visual clarity. A plot with its performance included is available in the Supporting Information (Figure~\ref{fig:nll_full}).

On all datasets, the MPNN GP, MPNN RF, FP RF and FP GP performed well. While median scores were strong for the MPNN MVE as well, this method and the FFN MVE showed by far the most variation across random splits. The MPNN Ensemble consistently underestimates error; this tendency was observed in the MPNN Bootstrap and MPNN Snapshot Ensembling as well, although to a lesser extent. All estimators performed far better on the CLogP random split than the scaffold split, as shown in Figure~\ref{fig:nll}e, suggesting that many estimators struggle to account for aleatoric uncertainty.

\subsubsection{Calibrated NLL}

\begin{figure}
    \centering
        \includegraphics[scale=0.5]{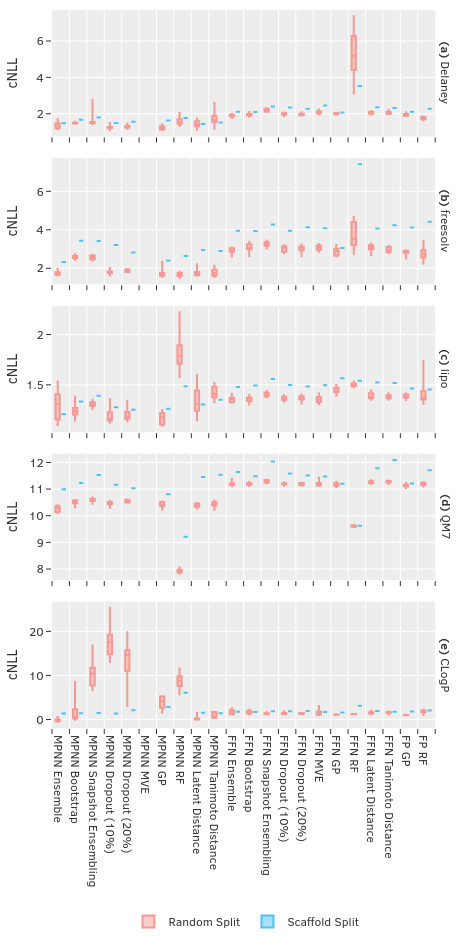}
    \caption{Calibrated negative log likelihood (cNLL) for all UQ methods (excluding MPNN MVE) and datasets evaluated in this work. A lower value of cNLL indicates better quantitative agreement between the estimated uncertainty and the magnitude of the observed absolute error after allowing a linear adjustment to the UQ metric to describe the predicted variance of errors.  (a) Delaney aqueous solubility, (b) freesolv solvation energy, (c) lipophilicity, (d) QM7 atomization energy, (e) CLogP heuristically-calculated lipophilicity. Random split boxes show quartiles without excluding  outliers.}
    \label{fig:cnll}
\end{figure}

The final set of results show the calibrated negative log likelihoods (cNLL) for each dataset and method (Figure~\ref{fig:cnll}). This is a more forgiving quantitative assessment, as each evaluation reflects an optimized calibration of $\hat \sigma^2(x) \coloneqq a U(x) + b$ as defined by Equation~\ref{eq:cnll}. MPNN MVE produced such extreme values on certain splits that it has been omitted for visual clarity. A plot with its performance included is available in the Supporting Information (Figure~\ref{fig:cnll_full}).

Calibrating each $U(x)$ has an equalizing effect and causes many methods to have similar cNLL values. It is essential for evaluation of relative UQ metrics like the Tanimoto or latent spaces distances and provides a quantitative boost to others. For many methods, however, the optimized parameters may lead to predicting a nearly-constant variance ($a \approx 0$) and so the calibrated model ignores structure-specific uncertainties. More information on this phenomenon can be found in the Supporting Information (Figure~\ref{fig:calibration_plot}).

On all the first four datasets, calibration is quite effective at improving some methods, reducing the maximum observed NLL across all estimators and splits by nearly an order of magnitude. However, calibration is sensitive to the choice of calibration set (here, the entire validation data) and harms the performance of those methods which were previously well calibrated, such as the MPNN MVE, MPNN GP, and MPNN RF. This is most evident in Figure~\ref{fig:cnll}e, where the performance of several methods appears significantly worse once calibrated. This effect may be due to particularly poor fits on certain random splits.

\subsection{Pairwise Model Comparisons}

\begin{figure}
    \centering
        \includegraphics[scale=0.25]{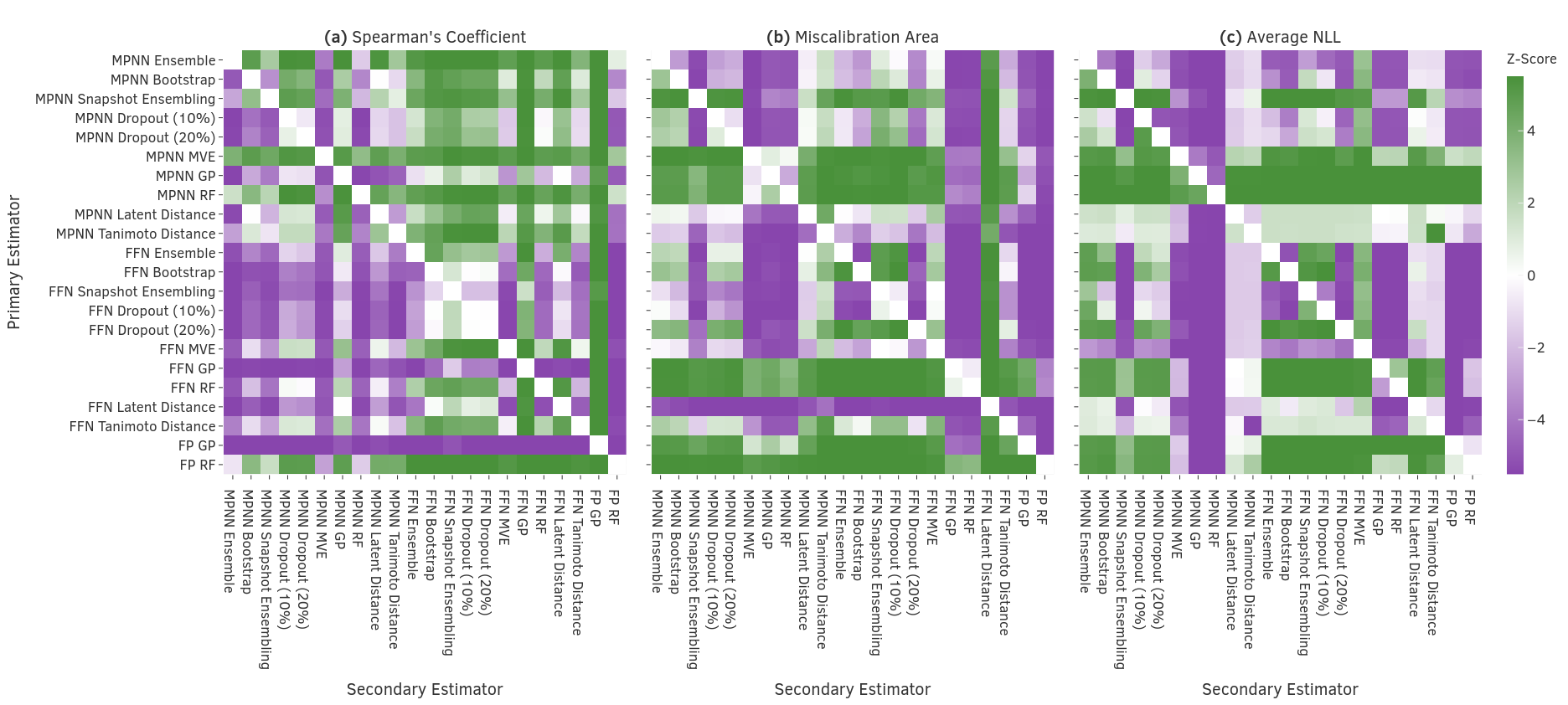}
    \caption{Z-scores produced by a modified Wilcoxon Signed-Rank Test (WSRT), computed over all data sets and random splits, for every pair of UQ methods when performance is ranked by (a) Spearman's Coefficient, (b) Miscalibration Area, and (c) NLL. A positive z-score indicates that the \textit{Primary Estimator} (y-axis) often ranked better than the \textit{Secondary Estimator} (x-axis) on that particular metric.}
    \label{fig:wilcoxon_splits}
\end{figure}

To provide an overview of each UQ method's relative performance across datasets, we perform modified Wilcoxon Signed-Rank Tests (WRSTs).\cite{wilcoxon1992individual, demsar_statistical_2006} Given a performance metric and two UQ methods (a \textit{primary} and \textit{secondary}), the WRST aims to determine how frequently 
the primary method outranks the secondary across all five datasets and eight random splits. A large positive z-score suggests that the primary method consistently outperformed the secondary method, and a large negative z-score suggests that it consistently underperformed. Figure~\ref{fig:wilcoxon_splits} shows the WSRT z-scores for every pair of UQ methods when ranked by Spearman's Coefficient, Miscalibration Area, and NLL. Rows with many positive z-scores (indicated by the color green in this plot) indicate UQ methods that consistently performed well when compared against any of the tested alternatives.

Note that the z-score is calculated purely on the ranked performance of the methods and does not indicate the extent to which one method outperformed another. 
More technical details regarding the calculation of WRST and additional figures are available in the Supporting Information, including a comparison of the median performance when the eight random splits are aggregated (Figure~\ref{fig:wilcoxon_splits_median}).

\section{Discussion}

Our analysis examines the performance of several approaches to UQ using learned embeddings via message passing networks (MPNNs) or feed forward networks (FFNs) using fixed fingerprint (FP) representations. Our evaluation focuses on four metrics: the Spearman's rank correlation coefficient, measuring relative agreement between UQ metrics and absolute errors; the miscalibration area, measuring overall agreement between predicted UQ metrics as predicted variances for normally-distributed errors; the average negative log likelihood of the observed value, using models' predicted values and UQs as means and variances of normal distributions; and a calibrated version of the negative log likelihood, where the UQ metrics are linearly related to variances.

\subsection{Consistency of performance across model architectures}
One challenge in comparing UQ method performance when applied to MPNNs and FFNs is that the models themselves have very different accuracies. As shown in Figure~\ref{fig:rmse}, RMSEs produced by FFNs were often twice as high as those produced by MPNNs on the same dataset. These results are consistent with the findings of \citeauthor{yang2019chemprop} \cite{yang2019chemprop}. However, the same figure also shows that MPNNs and FFNs usually saw a similar reduction in RMSE, as a percent of total, from each UQ method. All UQ methods also produced similar Spearman's rank correlations, miscalibration areas, and NLLs across model architectures. There was some more variation on Calibrated NLL, on which UQ methods appeared to perform uniformly better on MPNNs than FFNs, except on CLogP dataset, where the reverse was true. As discussed in the NLL results section, this can be explained by the significant gap in accuracy between the MPNNs and FFNs. For both MPNNs and FFNs, we observed union-based methods to  outperform virtually all ensembling methods in terms of miscalibration area and NLL (Figure~\ref{fig:wilcoxon_splits}bc).

\subsection{Consistency of performance across datasets}
Methods did not perform consistently relative to one another across different datasets, with some  achieving standout performance on certain datasets and middling performance on others. This variation explains why one proposed method may appear superior to other baselines for a particular  dataset or split. For example, \citeauthor{janet2019quantitative} found the calibrated Latent Distance UQ to significantly outperform uncalibrated ensembling on the two datasets they tested using a FFN with an application-informed feature vector\cite{janet2019quantitative}.

The variation observed in Spearman's rank correlation across datasets was enough to significantly affect practical utility of the UQ. For example, the MPNN MVE achieved an average Spearman's Rank Correlation of approximately $0.65$ on CLogP and approximately $0.1$ on lipo. This translated to much poorer isolation of low error predictions on the latter dataset: The top $25\%$ of predictions identified by the MPNN MVE on CLogP had an RMSE that was $75\%$ less than the RMSE on the entire test set, while the top $25\%$ of predictions using the MPNN MVE on lipo had \emph{higher} RMSE than the overall test set.

\subsection{Consistency of performance across evaluation metrics}
Method performance was also not consistent across evaluation metrics, although MPNN RF and MPNN MVE were highly ranked for each. The relevant subset of our results are consistent with those of \citeauthor{scalia2019evaluating}, who found that while  MPNN Ensemble generally produces a ranking at least as strong as MPNN Bootstrapping, the latter method is better calibrated.\cite{scalia2019evaluating}

As illustrated by Figure~\ref{fig:wilcoxon_splits}, we found that a ranking of uncertainties was best achieved using MPNN MVE or MPNN RF approaches, with MPNN Ensemble and FP RF not far behind. Overall miscalibration areas were lowest using FP RF, followed by FFN RF or FFN GP and MPNN GP, MPNN RF, or MPNN MVE. Negative log likelihoods, reflecting both quantitative accuracies and uncertainty estimates, were lowest using MPNN FP, MPNN GP, and MPNN MVE. 

\subsection{Highest-performing method}

Figure~\ref{fig:wilcoxon_splits}c shows  that MPNN RF very often produced the lowest NLL of all tested methods, with MPNN GP and MPNN MVE in second and third place; FP RF and FP GP were not far behind in fourth and fifth.  These results are consistent with those of \citeauthor{tran2019material}, who when testing UQ methods on the material property prediction space found their Convolution-Fed Gaussian Process---an implementation similar to our MPNN GP---to perform best.\cite{tran2019material} While we found the MPNN GP to perform worse on all three metrics than the MPNN RF, the differences were relatively minor and it might be possible to close the difference in ranking performance with hyper-parameter optimization. The MPNN RF, MPNN GP, and MPNN MVE methods were found to be the highest-performing neural network methods overall with FP RF representing a strong baseline.

\section{Conclusion}
An ideal uncertainty estimator would exhibit significantly less performance variation between datasets than has been observed from these existing methods. One technique which may lead to more consistently performant models is stacking, which involves joining the results of multiple weak models to produce a meta-estimator.\cite{breiman_stacked_1996} \citeauthor{scalia2019evaluating} previously experimented with summing the outputs of an MPNN MVE with a few ensemble based methods, but there remain many unexplored method combinations and aggregation strategies.\cite{scalia2019evaluating} 
If several methods were known with near-independent performance, there would be a high probability that at least one of them would work well for any particular task; the challenge would then be to learn when each estimator is most relevant. It will also be worthwhile to explore the performance of these methods, or approximations of these methods, on larger datasets, given that several of the implementations presented here (e.g., ensembling) scale impractically. Finally, future work might expand on recent work on UQ for classification tasks.\cite{yang_comprehensive_2020} Given the current state of the art in UQ for regression models and our results, particularly using the NLL evaluation metric to reflect both regression accuracy and UQ accuracy, we  recommend the MPNN RF as a reasonable first approach, while again noting the substantial variation across datasets and splits.

\begin{acknowledgement}

We thank the Machine Learning for Pharmaceutical Discovery and Synthesis consortium and the DARPA Accelerated Molecular Discovery program for funding this research. LH thanks Takeda for additional financial support through the Takeda Undergraduate Research and Innovation Scholarship. We thank members of the consortium for their helpful feedback throughout the research process. LH would also to thank José Cambronero, Patrick John Chia, and the SuperUROP staff for their advice and comments. The authors declare no competing financial interest. 

\end{acknowledgement}

\begin{suppinfo}

All code, data, and tabulated results used in this study can be found at \url{https://github.com/lhirschfeld/ChempropUncertaintyQuantification}. The Supporting Information contains additional results relating to the negative log likelihood evaluation and the Wilcoxon signed-rank test. %

\end{suppinfo}

\bibliography{main}

\newpage
\pagebreak

\renewcommand{\thefigure}{S\arabic{figure}}
\renewcommand{\thetable}{S\arabic{table}}
\setcounter{figure}{0} 
\setcounter{table}{0} 
\setcounter{page}{1}

\captionsetup{font=footnotesize}
\newgeometry{left=0.5in, right=0.5 in}

\begin{centering}

\textsf{\Large{\textbf{Supporting Information}}}

\vspace{0.5cm} 
\textsf{\Large{\textbf{Uncertainty Quantification in Molecular Property Prediction using Message Passing Networks}}}

\vspace{0.3cm}

\textsf{\large{Lior Hirschfeld$^\dagger$, Kyle Swanson$^\ddagger$, Kevin Yang$^\P$, Regina Barzilay$^{\ast,\dagger}$, Connor W. Coley$^{\ast,\S}$ }}\\
\begin{center}
\it  $\dagger$ Computer Science and Artificial Intelligence Laboratory, MIT, Cambridge, MA 02139\\
\it  $\ddagger$ Department of Pure Mathematics and Mathematical Statistics, University of Cambridge, Cambridge, UK CB3 0WB \\
\it  $\P$ Department of Electrical Engineering and Computer Sciences, University of California Berkeley, Berkeley, CA 94720 \\
\it  $\S$ Department of Chemical Engineering, MIT, Cambridge, MA 02139
 \\

\textsf{ E-mail: regina@csail.mit.edu; ccoley@mit.edu}
\end{center}

\end{centering}

\vspace{0.5cm}

\normalsize

\section{Additional Results}
\begin{figure}
    \centering
        \includegraphics[scale=0.5]{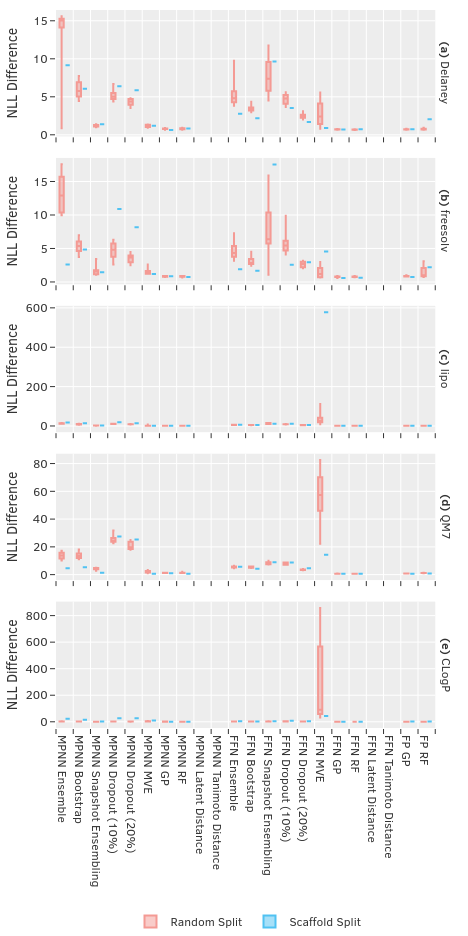}
    \caption{The difference between the observed negative log likelihood (NLL) and NLL that would be produced by an ideal estimator, for all UQ methods and datasets evaluated in this work. A lower value of NLL Difference indicates better quantitative agreement between the estimated uncertainty and the magnitude of the observed absolute error when treating the UQ metric as the predicted variance of errors. This evaluation interprets the UQ metric quantitatively, so we do not expect the distance-based metrics (in latent space or in structure space) to produce meaningful results. (a) Delaney aqueous solubility, (b) freesolv solvation energy, (c) lipophilicity, (d) QM7 atomization energy, (e) CLogP heuristicly-calculated lipophilicity. Random split box plots show quartiles without removing outliers.}
    \label{fig:nll_difference}
\end{figure}

\begin{figure}
    \centering
        \includegraphics[scale=0.5]{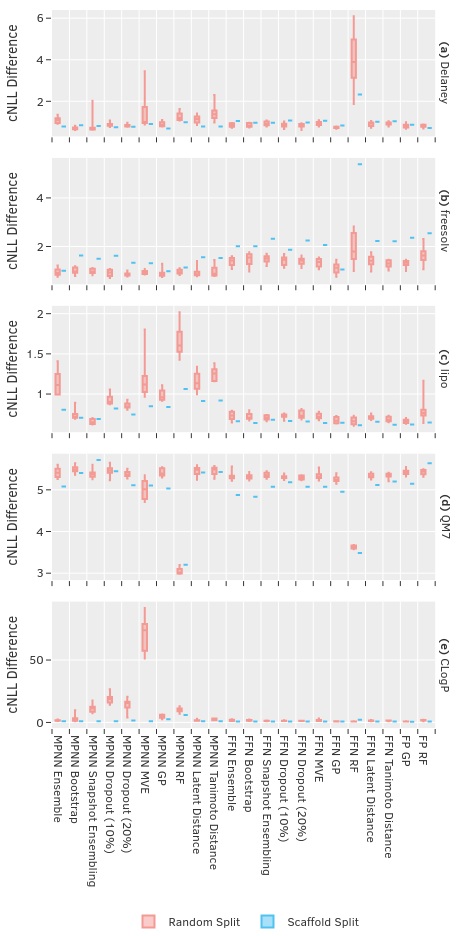}
    \caption{The difference between the calibrated negative log likelihood (cNLL) and the NLL that would be produced by an ideal estimator, for all UQ methods and datasets evaluated in this work. A lower value of cNLL Difference indicates better quantitative agreement between the estimated uncertainty and the magnitude of the observed absolute error after allowing a linear adjustment to the UQ metric to describe the predicted variance of errors.  (a) Delaney aqueous solubility, (b) freesolv solvation energy, (c) lipophilicity, (d) QM7 atomization energy, (e) CLogP heuristicly-calculated lipophilicity. Random split box plots show quartiles without removing outliers.}
    \label{fig:cnll_difference}
\end{figure}

\begin{figure}
    \centering
        \includegraphics[scale=0.5]{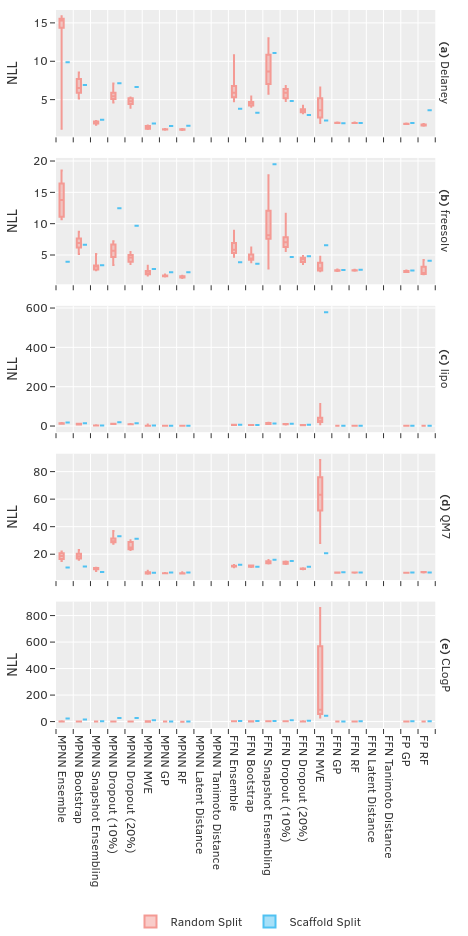}
    \caption{Negative log likelihood (NLL) for all UQ methods which produce confidence intervals and all datasets evaluated in this work. A lower value of NLL indicates better quantitative agreement between the estimated uncertainty and the magnitude of the observed absolute error when treating the UQ metric as the predicted variance of errors. This evaluation interprets the UQ metric quantitatively, so we do not expect the distance based metrics (in latent space or in structure space) to produce meaningful results. (a) Delaney aqueous solubility, (b) freesolv solvation energy, (c) lipophilicity, (d) QM7 atomization energy, (e) CLogP heuristically-calculated lipophilicity. Random split boxes show quartiles without excluding  outliers.}
    \label{fig:nll_full}
\end{figure}

\begin{figure}
    \centering
        \includegraphics[scale=0.5]{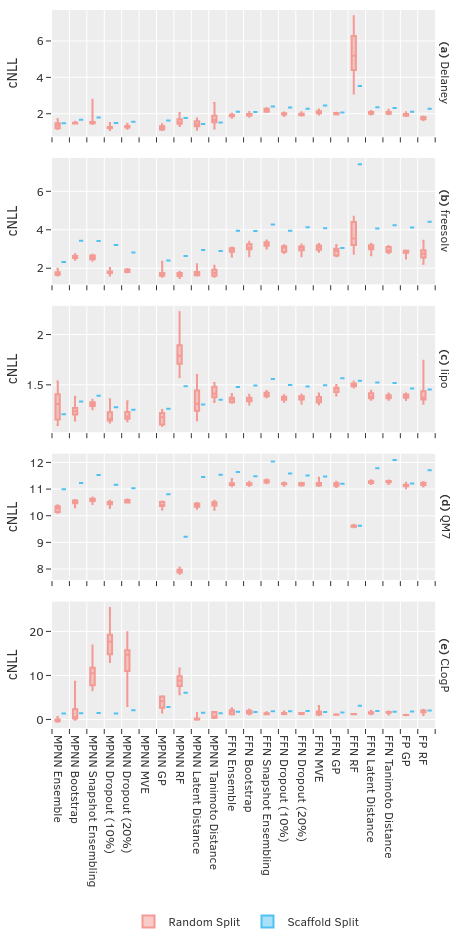}
    \caption{Calibrated negative log likelihood (cNLL) for all UQ methods (excluding MPNN MVE) and datasets evaluated in this work. A lower value of cNLL indicates better quantitative agreement between the estimated uncertainty and the magnitude of the observed absolute error after allowing a linear adjustment to the UQ metric to describe the predicted variance of errors.  (a) Delaney aqueous solubility, (b) freesolv solvation energy, (c) lipophilicity, (d) QM7 atomization energy, (e) CLogP heuristically-calculated lipophilicity. Random split boxes show quartiles without excluding  outliers.}
    \label{fig:cnll_full}
\end{figure}

\begin{figure}
    \centering
        \includegraphics[scale=0.5]{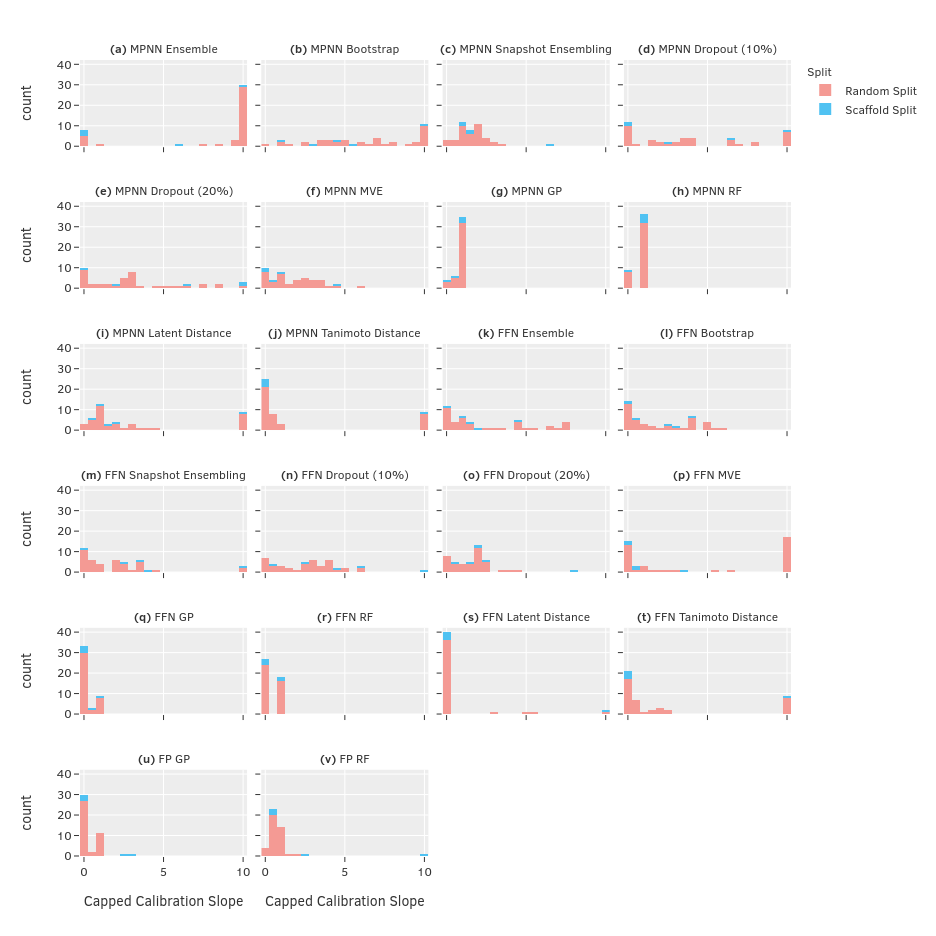}
    \caption{Calibration slopes for each method capped between $0$ and $10$, aggregated across data sets and splits. A high fraction of slopes above or below 1 indicates poor calibration. For example, the large number of slopes $\ge$ 10 for (a) MPNN Ensemble and (k) FFN Ensemble indicate that traditional ensembling consistently underestimates uncertainty.}
    \label{fig:calibration_plot}
\end{figure}

\clearpage
\section{Wilcoxon}
\label{sec:wilcoxon_supp}

\begin{figure}
    \centering
        \includegraphics[scale=0.25]{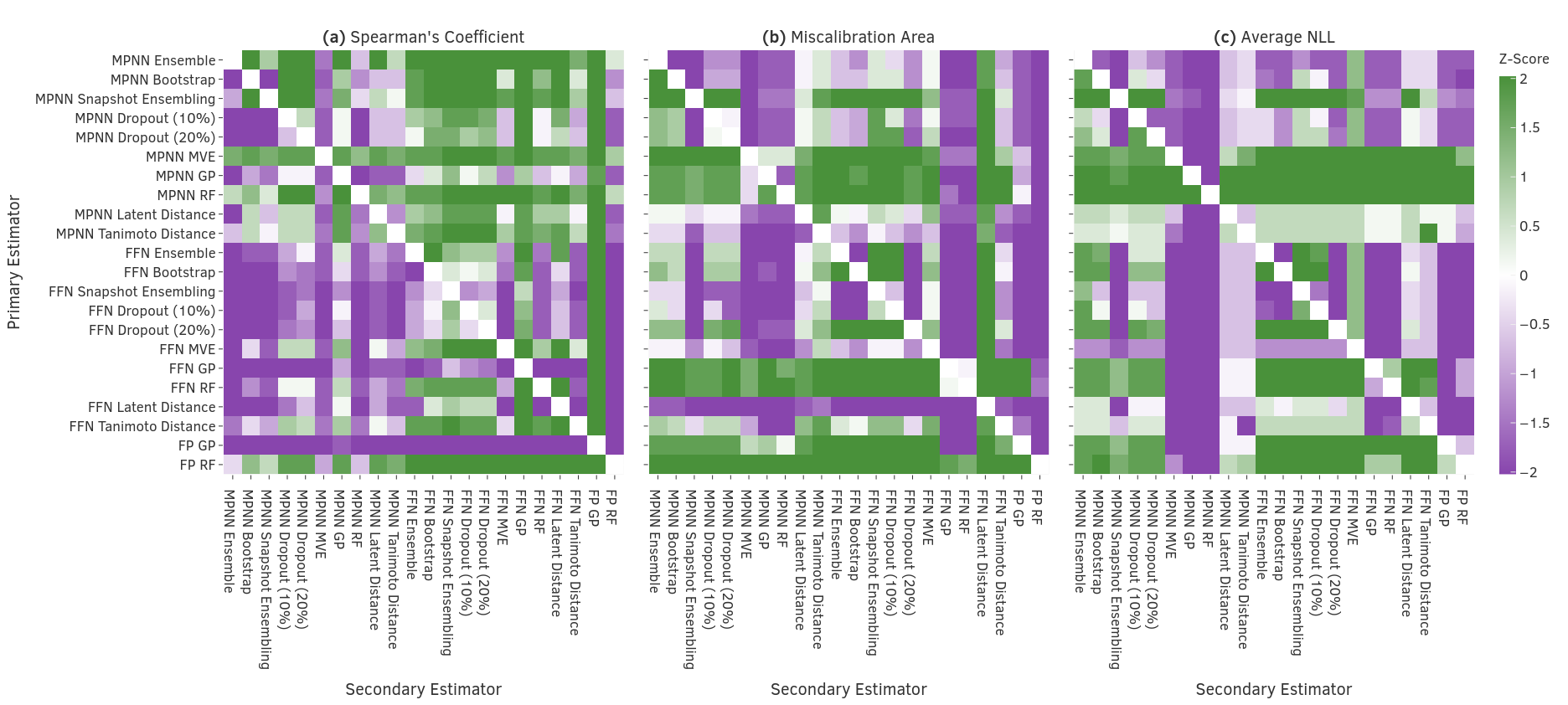}
    \caption{Z-scores produced by a modified Wilcoxon Signed-Rank Test (WSRT), computed over the median random split performance for each dataset, for every pair of UQ methods and three metrics considered in this work. A higher z-score indicates that the \textit{Primary Estimator} often ranked better than the \textit{Secondary Estimator} on that particular metric. (a) Spearman's Coefficient, (b) Miscalibration Area, (c) NLL.}
    \label{fig:wilcoxon_splits_median}
\end{figure}

Given a set of scores (performance metrics) for one UQ metric (the \textit{primary}), $\{x_1, x_2, \cdots, x_n\}$ and a set of scores for another UQ metric (the \textit{secondary}), $\{y_1, y_2, \cdots, y_n\}$ for $n$ different datasets/splits using one evaluation criterion, the WSRT z-score is calculated by first computing a set of differences $D = \{d_1, d_2, \cdots, d_n\}$, where

$$d_i = x_i - y_i$$.

The set is then sorted to produce $R$, where $r_i$ is the $i$th smallest element of $D$, by magnitude. Let $S = \sum _{i} i$ for all $i$ where $r_i$ indicated the primary primary UQ method outperformed the secondary (a positive value if large values of the metric are desired and a negative value if small values of the metric are desired).

The z-score is then

$$z = \frac{S - \frac{1}{4}n(n+1)}{\frac{1}{24}n(n+1)(2n+1)}.$$

This formula has been modified slightly from the standard WRST\cite{demsar_statistical_2006}  in order to produce a signed z-score that allows us to distinguish between cases where the primary metric performs significantly better than the secondary metric from cases where it performs significantly worse.

\end{document}